%% file: egpaper_camera_ready.tex
\documentclass[10pt,twocolumn,letterpaper]{article}

\usepackage{cvpr}
\usepackage{times}
\usepackage{epsfig}
\usepackage{graphicx}
\usepackage{amsmath}
\usepackage{amssymb}

\usepackage[svgnames,table]{xcolor} 
\usepackage{multirow}
\usepackage{float}
\definecolor{rgb1}{RGB}{214,  38, 40}   
\definecolor{rgb2}{RGB}{43, 160, 4}     
\definecolor{rgb3}{RGB}{158, 216, 229}  
\definecolor{rgb4}{RGB}{114, 158, 206}  
\definecolor{rgb5}{RGB}{204, 204, 91}   
\definecolor{rgb6}{RGB}{255, 186, 119}  
\definecolor{rgb7}{RGB}{147, 102, 188}  
\definecolor{rgb8}{RGB}{30, 119, 181}   
\definecolor{rgb9}{RGB}{160, 188, 33}   
\definecolor{rgb10}{RGB}{255, 127, 12}  
\definecolor{rgb11}{RGB}{196, 175, 214} 

\usepackage[breaklinks=true,bookmarks=false]{hyperref}

\cvprfinalcopy 

\def\cvprPaperID{3858} 

\ifcvprfinal\fi
\begin{document}

\title{RGBD Based Dimensional Decomposition Residual Network for 3D Semantic Scene Completion}


\author{Jie~Li$^{1,2 }$ 
\thanks{~First two authors contributed equally.}
\and Yu~Liu$^{2~*} $
\and Dong~Gong$^{2}$
\and Qinfeng~Shi$^{2}$
\and Xia~Yuan$^{1}$
\and Chunxia~Zhao$^{1}$
\and Ian~Reid$^{2}$
\\
$^{1}$Nanjing University of Science and Technology, China\\
$^{2}$The University of Adelaide, Australia\\
}

\maketitle
\thispagestyle{empty}

\begin{abstract}
RGB images differentiate from depth as they carry more details about the color and texture information, which can be utilized as a vital complement to depth for boosting the performance of 3D semantic scene completion (SSC). SSC is composed of 3D shape completion (SC) and semantic scene labeling while most of the existing approaches use depth as the sole input which causes the performance bottleneck. Moreover, the state-of-the-art methods employ 3D CNNs which have cumbersome networks and tremendous parameters. We introduce a light-weight Dimensional Decomposition Residual network (DDR) for 3D dense prediction tasks. The novel factorized convolution layer is effective for reducing the network parameters, and the proposed multi-scale fusion mechanism for depth and color image can improve the completion and segmentation accuracy simultaneously. Our method demonstrates excellent performance on two public datasets. Compared with the latest method SSCNet, we achieve 5.9\% gains in SC-IoU and 5.7\% gains in SSC-IOU, albeit with only 21\% network parameters and 16.6\% FLOPs employed compared with that of SSCNet.
\end{abstract}

\section{Introduction}
\vspace{-0.1cm}
We live in a 3D world where everything occupies part of the physical space under the view of the human perception system. Similarly, 3D scene understanding is of importance since it is a reflection about the real-world scenario. As one of the most vital fields in 3D scene understanding, Semantic Scene Completion (SSC) has verity of applications, including robotic navigation~\cite{gupta2013perceptual}, scene reconstruction~\cite{hays2007scene}, auto-driving~\cite{laugier2011probabilistic} \textit{etc}. However, due to the dimensional curse brought by 3D representation~\cite{wang2010learning} and the limited annotation datasets, the research field of SSC still step slowly in the past decades.

With the renaissance of deep learning~\cite{krizhevsky2012imagenet, gong2017motion, yan2019two} and a few large-scale datasets being made available ~\cite{lin2014microsoft,deng2009imagenet,song2017_SSCNet} in recent years.
The research activities of 3D shape processing thrive again in the computer vision community, injecting new possibilities and objectives for SSC as well introducing some unprecedented challenges.

Conventional methods usually utilize the hand-crafted features, such as voxel~\cite{kim20133d} and TSDF~\cite{izadi2011kinectfusion} to represent the 3D object shape, and make use of the graph model to infer the scene occupations and semantic labeling individually~\cite{gupta2015indoor,kim20133d}. The current state-of-the-art technique SSCNet~\cite{song2017_SSCNet}, instead uses an end-to-end 3D network to conduct the scene completion and category labeling simultaneously. Through combining the semantic and geometrical information implicitly via the network learning process, the two individual tasks can benefit from each other. 

Though remarkable gains in terms of scene completion and labeling accuracy have been achieved, the massive amount of parameters brought by the 3D representation make it computing-intensive. Moreover, another problem suffered in the existing SSC is the low-resolution representation~\cite{song2017_SSCNet,guo2018_VVNet}. In particular, due to the limitation of computation resources, both of the conventional and deep learning based methods sacrifice high-resolution to compromise an acceptable speed. 

On the other hand, most of existing methods solely use depth as input, which is struggled to differentiate objects from various categories. For example, a paper and a tablecloth on the same table can be easily distinguished by color or texture information. To sum up, depth and color image are different modalities captured by the sensor, they all provide us with what the scene looks like. The former gives us more sense about the object shape and distance, while the later transfers more information about the object texture and saliency. It is proved that both of the two modalities are helpful to boost the performance of SSC task~\cite{Garbade2018_twoStream}, although how to fuse them is still an unsolved problem.

To overcome the problems as mentioned above, we propose a light-weight semantic scene completion network, which utilizes both of the depth and RGB information. It formulates the 3D scene completion and labeling as a joint task and learns in an end-to-end way. The main contributions of this paper are three-fold:

\begin{itemize}
\vspace{-0.3cm}
\item Firstly, we propose the dimensional decomposition residual (DDR) blocks for 3D convolution, which dramatically reduces the model parameters without performance degradation.
\vspace{-0.3cm}
\item Secondly, 3D feature maps of RGB and depth are fused in multi-scale seamlessly, which enhances the network representation ability and boost the performance of SC and SSC tasks.
\vspace{-0.3cm}
\item Thirdly, the proposed end-to-end training network achieves state-of-the-art performance on NYU~\cite{silberman2012indoor} and NYUCAD~\cite{firman2016NYUCAD} datasets.
\vspace{-0.3cm}
\end{itemize}

The rest of this paper is organized as follows. Section~\ref{relatedworks} briefly summaries the related works, Section~\ref{methodology} introduces the methodology. Section~\ref{experiments} presents the experimental results.
Section~\ref{ablationstudy} analyses different parts of the proposed method.
Section~\ref{conclusion} summarizes our findings and concludes with future research interests. 

\input{figs/fig_NetworkStructure.tex}

\section{Related Works}
\label{relatedworks}
\subsection{3D Scene Completion and Semantic Labeling}
\vspace{-0.1cm}
As an important branch in 3D scene understanding, semantic scene completion (SSC) has many real-world applications and has received increasing attention recently with the support of deep learning~\cite{krizhevsky2012imagenet} and the large-scale annotated dataset~\cite{song2017_SSCNet}.

SSCNet~\cite{song2017_SSCNet} is the first one which formulates the shape completion and semantic labeling as a joint task and learns the task in an end-to-end way. TS3D~\cite{Garbade2018_twoStream} is based on SSCNet, and utilizes an additional network to incorporate the color information into the learning loop. Both of SSCNet and TS3D adopt truncated signed distance function (TSDF~\cite{izadi2011kinectfusion}) to encode the 3D volume, where every voxel stores the distance value \textit{d} to its closest surface, and the sign of the value indicates whether the voxel is in free space or occluded. However, TSDF is computationally intensive since it requires the calculation of the distance between the points on the surface and each point belong to the objects. Although with remarkable performance achieved, the 3D convolution representation results in a network that is computationally expensive with highly redundant parameters. 

\subsection{Computation-efficient Networks}
\vspace{-0.1cm}
As a milestone in deep learning architectures, ResNet~\cite{he2016deep} uses a residual block to prevent the performance degradation that occurs when network layers become deep. The extreme deep network leads to the state-of-the-art performance in many tasks including image classification~\cite{krizhevsky2012imagenet}, object detection~\cite{ren2017faster,redmon2016you,liu2019learning} and segmentation~\cite{chen2018deeplab,he2017mask}. However, this is very expensive concerning computation resource and heavy-burden~\cite{he2016deep,krizhevsky2012imagenet}. To cater to the appeal for real-world applications, there is a trend to tailor the heavy-burden networks to the light-weight network in recent years.

\textbf{Feature Representation}
Considering the redundant information contained in the 3D scene completion, the first spectrum of work try to model the scene with sparse feature representation. Specifically, OctNet~\cite{riegler2017octnet} and O-CNN~\cite{wang2017cnn} utilize the Octree-based CNN to represent the 3D object shape. PointNet~\cite{charles2017pointnet} and Kd-Networks~\cite{klokov2017escape} employ point clouds to indicate the occupation of the scene. Although saving the memory and computation, the neighbor pixels are usually mapped to the same voxel, which inevitably causes the detail missing for semantic labeling and scene understanding. 

\textbf{Group Convolution}
In recent two years, there are several popular light-weight networks have been proposed, include MobileNet~\cite{howard2017mobilenets,sandler2018mobilenetv2} and ShuffleNet~\cite{Zhang2017ShuffleNetAE}. In MobileNet, depth-wise convolutions and point-wise convolutions are utilized to separate the channels as well as reduce the parameters and the calculations. In ShuffleNet, besides the group point-wise convolution and depth-wise convolution are adopted, shuffle layer is developed for information exchange between different shuffle units. However, the above models heavily rely on depth-wise convolution and group convolution, and mainly target at 2D networks thus can not directly be applied for 3D tasks.

\textbf{Spatial Group Convolution}
To improve the computing efficiency of the 3D network. 
EsscNet~\cite{zhang2018efficient} is introduced, rather than to conduct the group convolution on feature channel dimension, which adopts the group convolution on the spatial aspect. 
The drawback of spatial group convolution is that it splits the features manually into separate parts, which cause the performance drops. Meanwhile, the splitting process involves hash table maintaining and coordinate with other blocks, and is cumbersome for transplantation. On the contrary, the proposed DDR block is much flexible, and it can be planted to any network which contains the 3D modules.

\subsection{Modality Fusion in SSC}
\vspace{-0.1cm}
There are many works focused on RGBD fusion in 2D applications~\cite{wang2016learning, park2017rdfnet,chang2017matterport3d,qi20173d,gupta2015indoor}. 
RGBD sensor can capture the depth and color images simultaneously, although depth can be used to infer the geometry of the scene, which is too sparse to reconstruct the occluded parts of the scene. 
Compared with depth, color image carries more cues about texture, color, and reflections, which can be viewed as an essential complement to the depth for SSC task. 
Following the design philosophy of SSCNet, TS3D~\cite{Garbade2018_twoStream} adds the color image into the work-flow. However, the scene labeling needs to be estimated twice, and the depth flow and color flow are still apart from each other from the essential.

In ~\cite{guedes2018semantic}, two fusion strategies were proposed, one is early-stage fusion which concatenates the feature at the first layer, and another is mid-level fusion which concatenates the features before the output layer. Although follow the overall design and reuse the features of SSCNet, the performance of adopting both fusion strategies are unexpectedly worse than that of SSCNet.

The most related work for feature fusion is RDFNet~\cite{park2017rdfnet}, which utilizes multi-scale fused features from color images, and aims to build a 2D segmentation framework. However, fusing the features in the 3D network is much more challenging as mentioned before. In this paper, we propose a novel fusion strategy which effectively fuses the 3D depth and color features on multi-scales without bringing in extra parameters.  

\section{Methodology}
\label{methodology}
\subsection{Overview}
\vspace{-0.1cm}
This section presents the proposed light-weight network for SSC. 
The computation-efficient Dimensional Decomposition Residual (DDR) block, as well as a novel modality fusion module, are emphasized.
On the one hand, through dimensional splitting on 3D convolutions and dense connection, using DDR blocks can dramatically reduce the network parameters. On the other hand, through fusing the 3D features of depth and color image seamlessly, the proposed network can efficiently make use of the information captured by the RGBD sensors, and various modulates of inputs complement with each other thus boost the performance of shape completion and scene labeling simultaneously.  
The framework of the proposed network is shown in Figure~\ref{fig:network_structure}. The network has two feature extractors, which take a full resolution depth and the corresponding color image as inputs, respectively. The network first uses 2D DDR blocks to learn the local textures and the geometry representation. Then, the 2D feature maps are projected to 3D space by a projection layer. A multi-level fusion strategy is then applied to fuse the texture and geometry information.
After that, the network responses are then concatenated and fed into the subsequent light-weight Atrous Spatial Pyramid Pooling (ASPP) module to aggregate information in multiple scales. In the end, another three pointwise convolutional layers are used to predict the final voxel labels. The following parts will explain the design details of each module.

\input{figs/ddrblocks.tex}

\subsection{Dimensional Decomposition Residual Blocks}
\vspace{-0.1cm}
\subsubsection{Basic DDR}
\vspace{-0.1cm}
Residual layers~\cite{he2016deep} have the property of allowing convolutional layers to approximate residual functions,
\begin{equation}\label{Eq:ResBlock2d_1}
x_t=\mathcal{F}^{d} \left(x_{t-1},\{ { W }_{ i }\}  \right) +x_{t-1}
\end{equation}
where $x_{t-1}$ and $x_{t}$ are the input and output.. 
The function $\mathcal{F}^{d} \left( x_{t-1},\{ { W }_{ i }\}  \right)$ represents the residual mapping to be learned and $d$ is the dilation rate within the block. 
This residual formulation facilitates learning and alleviates the degradation problem present in architectures that stack a large number of layers~\cite{romera2018erfnet}. 

Directly applying the original (2D) ResNet block into the 3D dense prediction task, the two corresponding 3D residual layers will be: the non-bottleneck design with two-layer $3\times 3\times 3$ convolutions as described  in Figure~\ref{fig:resddrblocks}(a), and the three-layer bottleneck version as depicted in Figure~\ref{fig:resddrblocks}(c). 

However, both of the two structures will suffer the problem of high computational costs as the network parameters grow in cubic. 
We propose to redesign the residual through decomposing the 3D convolution into three consecutive layers along each dimension. The proposed basic DDR block is shown in Figure~\ref{fig:resddrblocks}(b) and its deeper bottleneck version is shown in Figure~\ref{fig:resddrblocks}(d). In this way, the network can reduce parameters and capable of capturing 3D geometric information according to the theory in~\cite{szegedy2016rethinking}.

Here we provide an episode to illustrate the effectiveness of DDR block for reducing network parameters: 
Considering a 3D CNN with input channels $c^{in}$, output channels $c^{out}$, and kernel size of ${ k }^{ x }\times { k }^{ y }\times { k }^{ z }$. 
Without losing the generality, we can assume ${ k }^{ x }={ k }^{ y }={ k }^{ z }=k$. 
The original block in 3D CNN is then be decomposed into three consecutive layers with filter size $1\times 1\times k$, $1\times k\times 1$ and $k\times 1\times 1$, accordingly. The computational costs of the original block and DDR block are proportional to $c^{in}\times c^{out}\times k\times k \times k$ and $c^{in}\times c^{out}\times (k + k + k)$, respectively. The advantage of DDR for reducing network parameters will be enlarged when \textit{k} become large, since $3k\ll { k }^{ 3 }$. 
As an example, the parameters of a typical 3D convolutional layer with a $3\times 3\times 3$ kernel will drops to 1/3 after adopting DDR block.

\vspace{-0.4cm}
\subsubsection{Deeper DDR}
\vspace{-0.1cm}
Inspired by the bottleneck design~\cite{he2016deep}, we further deliver a deeper DDR block. In specific, for each residual function, a $1\times 1\times 1$ layer is added at both the top and bottom of the Dimensional Decomposition convolutions. The $1\times 1\times 1$ layers are responsible for reducing and restoring the dimensions, which make the three Dimensional Decomposition convolutional layers form a bottleneck with smaller input/output dimensions.

Moreover, parameter-free identity shortcuts are added within each dimensional decomposition convolution. 
The dense identity connections are not only helpful for robust feature representation but also have the merits of alleviating the vanishing-gradient problem and strengthen the feature propagation~\cite{he2016deep}.
In the remainder of the paper, DDR refers to the deeper DDR block unless specifically noted.

\subsection{Two Modality Multi-level Feature Fusion}
\vspace{-0.1cm}
\subsubsection{Feature Extractor Module}
\vspace{-0.1cm}
In our network, there are two parallel branches for feature extraction corresponding to the depth and color image. As shown in Figure~\ref{fig:network_structure}(b), the feature extractor module is composed of three components: a 2D feature extractor, a 3D feature extractor and a projection layer which mapping the 2D feature to the 3D feature. The network first utilizes 2D feature extractor to learn local color and texture representation. After feature mapping to the 3D space by a projection layer, 3D feature extractor is employed to acquire the geometry and context information.

\textbf{2D Feature Extractor}
To extract features from a 2D depth and color image, a 2D point-wise convolution is firstly used to increase the channels of feature maps. Then two 2D DDR blocks are stacked for residual learning. Through the process, the resolution of the output feature map keeps the same as the input image.
Please note, in our network, the number of parameters for 2D DDR blocks is 192, which is insignificant when compared with the 195k parameters for 3D DDR blocks. Therefore, we mainly focus on the light-weight operations of 3D DDR blocks. 

\input{figs/resNYU.tex}

\textbf{Projection Layer}
Since each pixel in the depth map corresponding to a tensor in the 2D feature map, every feature tensor can be projected into the 3D volume at the location with the same depth value. This step yields an incomplete 3D volume that assigns to every surface voxel its corresponding feature tensor. For the voxels that are not occupied by any depth values, their feature vectors are set to zeros. The mapping index  ${\mathcal{T}}_{u,v}$ at $(u,v)$ can be computed using the depth value $I_{u,v}$ and camera pose $C$ which are provided along with each image. Because the feature volume resolution is lower than the feature map resolution, several neighboring features will be projected into the same voxel, and we use max-pooling to simulate this step. With the feature projection layer, the 2D feature maps extracted by the 2D CNN are converted to a view-independent 3D feature volume. During training, the mapping indexes ${\mathcal{T}}$ between feature map tensors and voxels are recorded in a table for gradient back-propagation.

\textbf{3D Feature Extractor}
After the feature projection layer, a view-independent 3D feature volume is acquired. In this step, we further extract features using two 3D DDR blocks. A down-sample block is added in front of each DDR blocks to reduce the size of the feature maps and increase the dimension of its channel. Figure~\ref{fig:network_structure}(c) shows the structure of the down-sample block. A pooling layer and a pointwise convolution layer are concatenated to increase the channels of the output feature map of the down-sample block.

\subsubsection{Multi-level Feature Fusion}
\vspace{-0.1cm}
One primary challenge of 3D RGBD based semantic segmentation is how to effectively extract the color features along with depth features and to utilize those features for the labeling. To fully use the multi-modal features, we propose a novel feature fusion strategy which is inspired by ~\cite{lin2017refinenet,park2017rdfnet}. We employ multi-modal CNN feature fusion while preserving the lower computational cost. In specific, different levels of features are extracted through multiple DDR modules,  and then these features are merged together by element-wise add. The reason for using element-wise add rather than other operations is because it can fuse the features neatly with insignificant computation costs.

Through the cascaded DDR blocks, both low-level features and high-level features are captured, which enhance the representation ability of the network and is beneficial for the performance of semantic scene completion task.

\input{figs/resNYUCAD.tex}

\subsection{Light-weight ASPP Module}
\vspace{-0.1cm}
Different object categories have various physical 3D sizes in indoor scenes. This requires the network to capture information at different scales in order to recognize the objects reliably. Atrous spatial pyramid pooling (ASPP)~\cite{chen2018deeplab,chen2018encoder} exploits multi-scale features by employing multiple parallel filters with different dilatation rates and has been proved to be powerful to improve the CNN's ability to handle objects with various sizes. However, directly applying ASPP in 3D semantic scene completion would bring in tremendous parameters as well as large computations.

Based on this consideration, we introduce a light-weight ASPP (LW-ASPP) which is capable of handling scale variability while requiring fewer computations. In specific, LW-ASPP uses multiple parallel DDR blocks with different sampling (dilation) rates. The dilated DDR is implemented by setting a dilation rate in the three-dimensional decomposition convolutions within the DDR block. The dilated DDR explicitly adjusts the field-of-view of filters as well as controls the resolution of the feature responses. The features extracted from different sampling rates are further concatenated and fused to generate the final result with the output layer, which is constructed by the three 3D point-wise convolution layers as shown in Figure~\ref{fig:network_structure}.

\input{figs/visRes.tex}

\subsection{Training and Loss}
\vspace{-0.1cm}
\noindent
\textbf{Training}
Given the training dataset (\ie the RGBD images and ground truth volumetric object labels of 3D scenes), our method can be trained end-to-end. SSCNet~\cite{song2017_SSCNet} sets a small value (0.05) as the weight of the voxels in free space for data balancing in the training process. We adopt the same strategy in our early training process. 
With each additional 50 training epochs, the weight of empty voxels is gradually doubled until it is set to be the same as the other occupied voxels.
All the experiments are conducted using the pyTorch framework on GPU. Our model is trained using the SGD optimizer with a momentum of 0.9, weight decay of $10^{-4}$ and batch size is 2, the initial learning rate is 0.01 and divided by a factor of 10 when the training loss changes less than 1e-4 within 5 consecutive epochs.

\textbf{Loss}
For training the network, we employ the softmax cross entropy loss on the unnormalized network outputs $y$:
\begin{equation}\label{Eq:loss}
\mathcal{L} =-\sum _{ c=1 }^{ N }{ { w }_{ c }{ \hat { y }  }_{ i,c }\log { \left( \frac { { e }^{ { y }_{ ic } } }{ \sum _{ c' }^{ N }{ { e }^{ { y }_{ ic' } } }  }  \right)  }  } 
\end{equation}
where ${ \hat { y }  }_{ i,c }$ are the binary ground truth vectors, \ie ${ \hat { y }  }_{ i,c } = 1$ if voxel $i$ is labeled by class $c$, $N$ is the number of classes, and ${ w }_{ c }$ is the loss weight. To compute the loss function, we remove all voxels outside the field of view and the room and include all non-empty voxels plus occluded voxels.

\input{figs/rgbdepth.tex}

\section{Experiments}
\label{experiments}
\vspace{-0.1cm}
In this section, we evaluate and compare the proposed method with the state-of-the-art approaches on two public datasets, \ie NYU~\cite{silberman2012indoor} and NYUCAD~\cite{firman2016NYUCAD}.
Both the quantitative and qualitative results demonstrate the superiority of our algorithm on SSC task. 

\subsection{Dataset and Metrics}
\vspace{-0.1cm}
\noindent
\textbf{Dataset}
We evaluate the proposed method on the NYUv2 dataset~\cite{silberman2012indoor}, which is in the following denoted as NYU. 
NYU consists of 1449 indoor scenes that are captured via a Kinect sensor. Following SSCNet~\cite{song2017_SSCNet}, we use the 3D annotated labels provided by~\cite{rock2015completing} for semantic scene completion task. 
NYUCAD~\cite{firman2016NYUCAD} uses the depth maps generated from the projections of the 3D annotations to reduce the misalignment of depths and the annotations.
We compare our method with the state-of-the-art methods on both NYU and NYUCAD datasets.

\noindent
\textbf{Metrics}
As the evaluation metric, the voxel-level intersection over union (IoU) between the predicted voxel
label and ground truth label is used. 
For the task of semantic scene completion, we evaluate the IoU of each object classes on both the observed and occluded voxels. For the task of scene completion, we treat all non-empty object class as one category and evaluate IoU of the binary predictions on the occluded voxels.

\subsection{Comparisons with the State-of-the-art Methods}
\vspace{-0.1cm}

Table~\ref{tab:QuantitativeResultsonNYU} shows the results on NYU dataset acquired by our method and the state-of-the-art methods. We achieve state-of-the-art performance regarding different metrics. Specifically, we achieve the best performance for both the tasks of scene completion and semantic scene completion and also rank the second best of recall and precision for scene completion. We outperform the previous SSCNet by a significant margin in overall performance, that are 5.7\% gains in semantic scene completion and 5.9\% gains in scene completion. The proposed network demonstrate the superior performance in some categories such as \textit{ceil.}, \textit{table}, \textit{tvs}, \textit{furn.} \etc. We inspect this improvement due to the novel architecture, that makes use of the robust features from multi-level and multi-modalities, and data fusion, which effectively complement the details from the color image to these textureless objects.

To validate the robustness and generalization of the proposed network, we also conduct experiments on NYUCAD dataset as shown in Table~\ref{tab:QuantitativeResultsonNYUCAD}. The comparison results with the state-of-the-art methods present the same trend. Among all of the methods, we achieve the best performance for semantic scene completion and scene completion.

\subsection{Quantitative Analysis}
\vspace{-0.1cm}
Since we target at a light-weight 3D network for semantic scene completion, in this section, we list the params and FLOPs of the proposed method as well as the baseline method. As shown in Table~\ref{tab:ParamsFLOPsofDifferent}. In specific, compared with the state-of-the-art method SSCNet, the parameters in our method is 21.0\% of that in SSCNet, and the FLOPs is 16.6\% of that SSCNet. However, the performance of both scene completion and semantic scene completion is around 6\%  higher than that of SSCNet. Compared with the EsscNet~\cite{zhang2018efficient}, depth solely is used as the input for a fair comparison, our method is computationally cheaper than EsscNet with 6\% reduction in FLOPS and increased performance. For SC and SSC tasks, EsscNet reaches the accuracies of 56.2\% (SC) and 26.7\% (SSC), and we achieve 59.0\% (SC) and 28.9\% (SSC).

\input{figs/resQuantitative.tex}

\subsection{Qualitative Analysis}
\label{qualitativeanalysis}
\vspace{-0.1cm}
Figure~\ref{fig:viz} shows visualized results (in different scenarios) of the scene segmentation generated by the proposed method (c) and SSCNet (d), ground truth (b) are also provided as a reference. All the results are acquired on the NYUCAD validation set. As can be seen, compared with SSCNet, the scene completion results of our method is much more abundant in detail and less error-prone.

It can be easily seen that our method performs better for objects such as \textit{furn}, \textit{wall}, and \textit{floor}. For example, in the second and third rows, SSC will cause some missing in the details of the wall, which is rarely happening in our algorithm. Part of the reason that our method is better for handling the texture-less and small objects we attribute which come from the incorporated color features.
In row(1), our method effectively captures the detail information about the leg of the chair. In addition, compared with SSCNet, the proposed method maintains the segmentation consistency for objects with big sizes, such as the \textit{wall} and \textit{floor} in the row(2) and \textit{ceiling} in the row(4). And row(3) shows a much challenging instance, \ie the window, both SSCNet and our method cannot acquire satisfied results. However, our method can recognize part of the information. Row(5) and row(6) show the failure cases in our methods, specifically, in the row(5), the \textit{fresco} on the wall has the similar texture with the stuff on the bookshelf, it thus wrongly classified into the category of the \textit{bookshelf}. In row(6), the ground-truth \textit{furniture} circled by the red dashed rectangle, SSCNet wrongly predicts it into the object category, and our network wrongly classifies it as a chair, which may due to the quite similar shape and color information between the furniture and the chair category. In supplementary materials, more visualized results are provided. 

\section{Ablation Study}
\label{ablationstudy}
\vspace{-0.1cm}
\noindent
\textbf{RGB and Depth Fusion}
Both RGB and Depth information are important for  3D scene understanding. To verify the effectiveness of the proposed multi-level fusion strategy, we evaluate the performance of our method with only depth or RGB image as the input. 
As can be seen in Table~\ref{tab:ablationrgbdepth}, and the performance on SSC of our method for only using depth or color image as input are 28.9\% and 27.2\%, respectively. Since RGB images carry more details such as color and texture, which is beneficial for the semantic information,  this can be seen from the results of category \textit{tvs} and category \textit{sofa}. However, the advantage of using depth lies on it carries more geometry information, for the objects which are difficult to differentiate through color information, it is much easier to tell the difference according to their shapes. Such as \textit{table} and \textit{floor}. Moreover, depth is less sensitive regarding illumination changes and the dramatic color variation within the same category, which may explain for the indoor scene, the result of using depth is a bit better than that of using a color image as input.

Meanwhile, merging depth and color features in our method significantly improve the SSC performance, which proves the two-modality information can be an excellent complement to each other. And benefit from the light-weight DDR block applied in the network, the overall computations and parameters remain small.

\input{figs/paramsASPP.tex}

\noindent
\textbf{Light-weight ASPP}
The effectiveness of ASPP has been verified in 2D semantic segmentation~\cite{chen2018deeplab} task. However, the direct expansion of ASPP from 2D to 3D would bring in a massive amount of parameters as well as make the network cumbersome. Lightweight ASPP (LW-ASPP) using DDR block as the primary core, which not only effectively reduces the network parameters but also inherits the merits of ASPP for capturing multi-scale information, thus is beneficial to the 3D task.

In order to verify the validity of LW-ASPP, we design a group of experiments in which LW-ASPP was removed from the network or replaced with 3D ASSP directly extended from ASPP. As can be seen in Table~\ref{tab:paramsASPP}, when compared with the network without ASPP module, adding LW-ASPP boosts the SC-IoU 3.2\% and SSC-IoU 3.6\%. When replacing LW-ASPP with 3D-ASPP, the performance can be further improved by a small margin but with the sacrifice of over two times params and around three times FLOPs. 

\noindent
\textbf{Change of speed/memory and performance}
As shown in Table~\ref{tab:SpeedMemory}, DDRNet with quite a few parameters and FLOPs compared to SSCNet. DDRNet has a deeper structure, thus stronger non-linear representation ability than the 3D-ResNet version, albeit less memory cost required. Moreover, DDRNet achieves much faster speed with an insignificant performance loss. 
\input{figs/tab_MemSpeed.tex}
\section{Conclusion}
\label{conclusion}
\vspace{-0.1cm}
This paper proposes a novel structure for handling the semantic scene completion problem. Specifically, an end-to-end light-weight Dimensional Decomposition Residual (DDR) network is delivered for scene completion and semantic scene labeling. The two contributions are the proposed factorized convolution layer and a novel two-modality fusion mechanism. The former is effective to reduce the parameters within the network, and the later can fuse the depth and color image seamlessly in multi-level, the state-of-the-art results are achieved for both SSC and SC task on two public datasets. In the future, considering to differentiate instances of the indoor scene as well as to incorporate the shuffle layer into the proposed light-weight network will be our research interests.

\section*{Acknowledgement}
\vspace{-0.1cm}
This work is supported by the National Natural Science Foundation of China under grants 61603184 and 61773210. We also gratefully acknowledge the support of the Australian Research Council through grants CE140100016 and FL130100102.


{\small
\bibliographystyle{ieee}
\bibliography{egbib}
}

\end{document}

%% file: figs/fig_NetworkStructure.tex


\begin{figure*}[t]
\begin{center}
{
\includegraphics[width=0.99\linewidth]{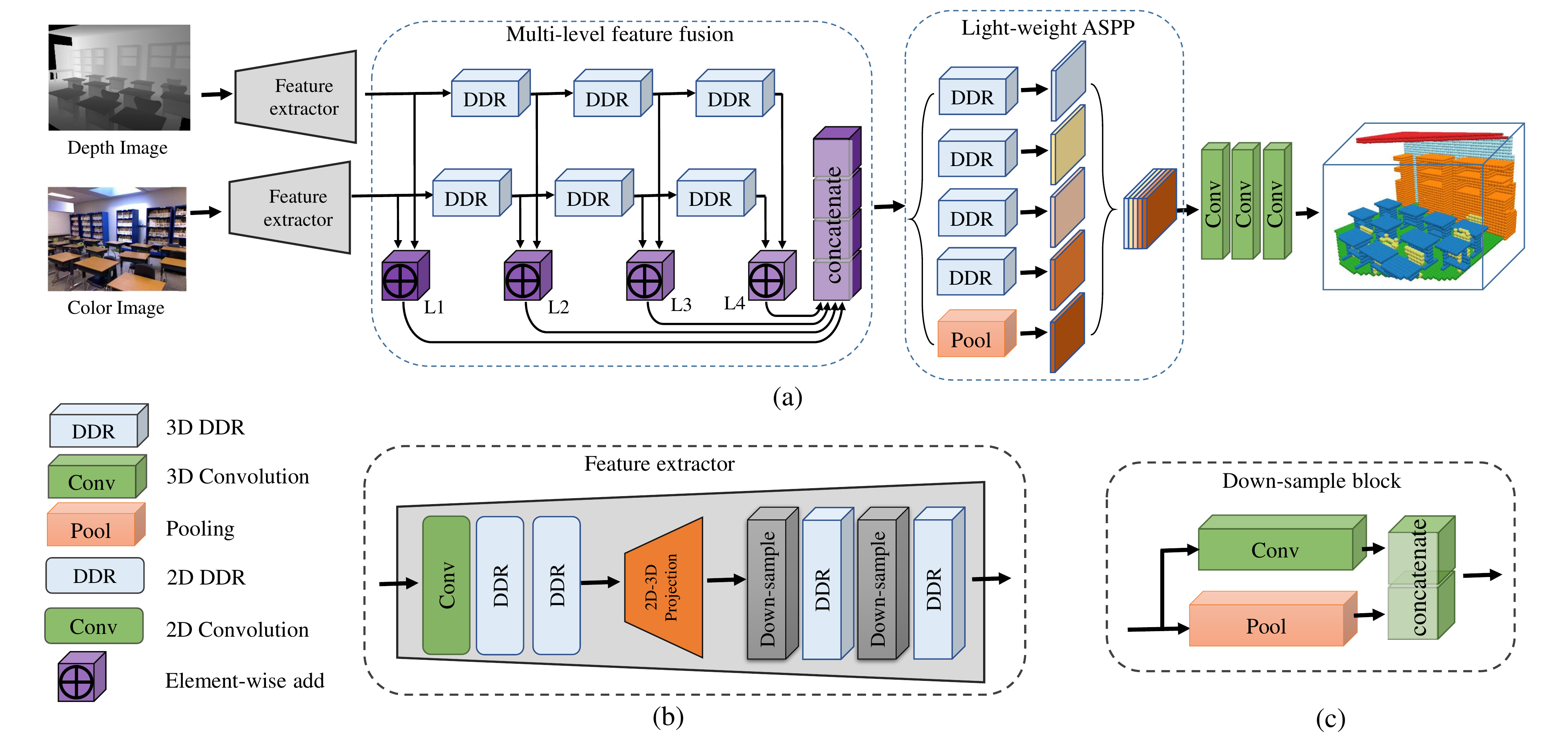}
}
\caption{(a) Network architecture for semantic scene completion. Taking RGBD image as input, the network predicts occupancies and object labels simultaneously. (b) Detailed structure of the feature extractor. (c) Structure of the down-sample block.}
\label{fig:network_structure}
\vspace{-0.6cm}
\end{center}
\end{figure*}

%% file: figs/ddrblocks.tex
\begin{figure}[t]
\begin{center}   
{
\includegraphics[width=0.8\linewidth]{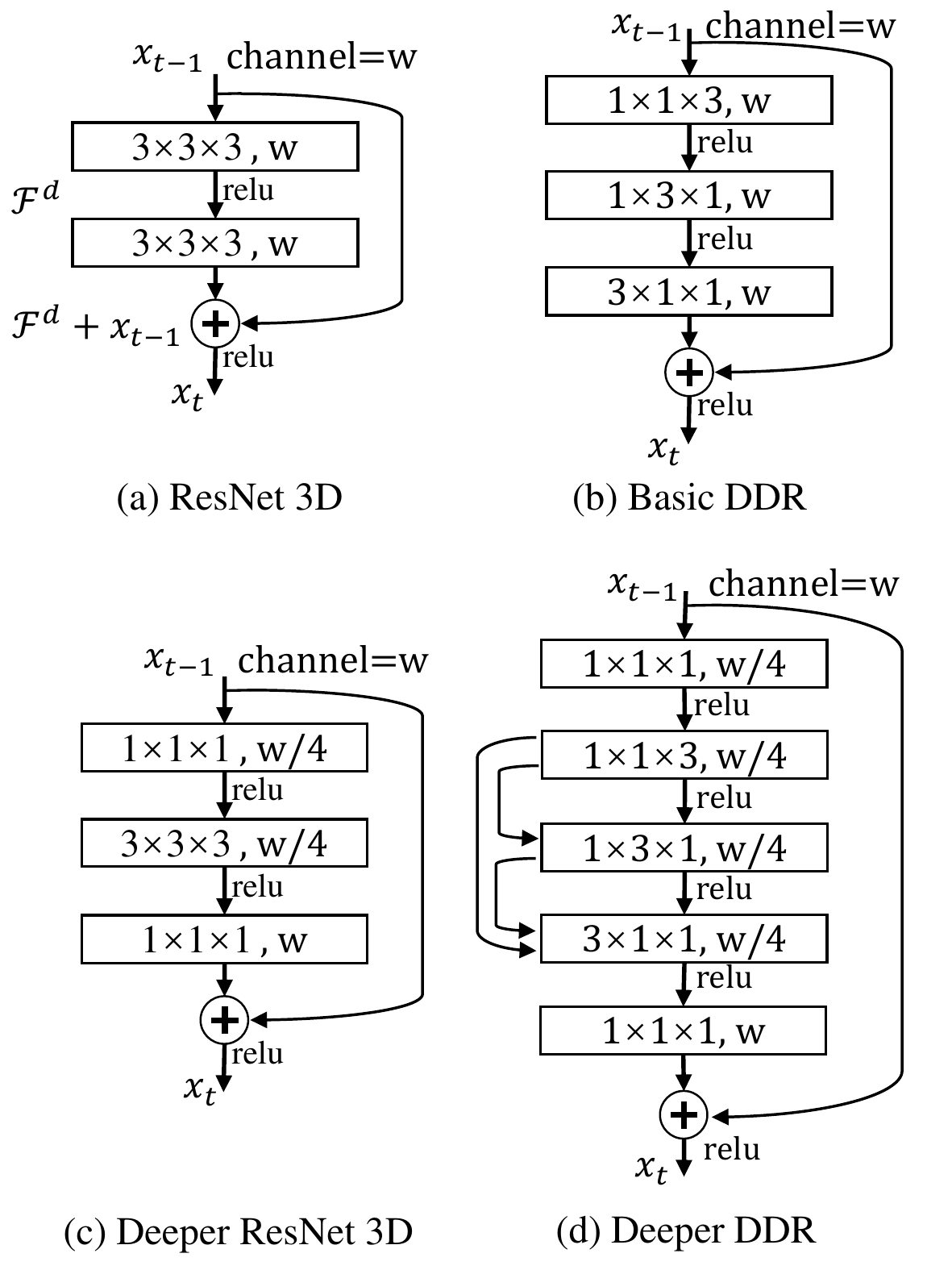}
}
\vspace{-0.2cm}
\caption{Residual blocks and the proposed DDR blocks.}
\vspace{-0.8cm}
\label{fig:resddrblocks}
\end{center}
\end{figure}


%% file: figs/resNYU.tex
\begin{table*}[t]
\begin{center}
\scalebox{0.9}
{
\begin{tabular}{l |c c c|c c c c c c c c c c c c} 
\hline
  & \multicolumn{3}{c|}{scene completion} & \multicolumn{12}{c}{semantic scene completion} \\ \hline
Methods  & prec. & recall & IoU & \cellcolor{rgb1}ceil. & \cellcolor{rgb2}floor & \cellcolor{rgb3}wall & \cellcolor{rgb4}win. & \cellcolor{rgb5}chair & \cellcolor{rgb6}bed & \cellcolor{rgb7}sofa & \cellcolor{rgb8}table & \cellcolor{rgb9}tvs & \cellcolor{rgb10}furn. & \cellcolor{rgb11}objs. & avg. \\ 
\hline
Lin \etal~\cite{lin2013holistic}   & 58.5 & 49.9 & 36.4 &  0.0 & 11.7 & 13.3 & {\bfseries 14.1} &  9.4 & 29.0 & 24.0 &  6.0 &  7.0 & 16.2 &  1.1 & 12.0\\
Geiger \etal~\cite{geiger2015joint} & 65.7 & 58.0 & 44.4 & 10.2 & 62.5 & 19.1 &  5.8 &  8.5 & 40.6 & 27.7 &  7.0 &  6.0 & 22.6 &  5.9 & 19.6\\ 
SSCNet~\cite{song2017_SSCNet} & 57.0 & {\bfseries 94.5} & 55.1 & 15.1 & {\bfseries 94.7} & 24.4 &  0.0 & 12.6 & 32.1 & 35.0 & 13.0 &  7.8 & 27.1 & 10.1 & 24.7\\
EsscNet~\cite{zhang2018efficient}   & {\bfseries 71.9} & 71.9 & 56.2 & 17.5 & 75.4 & 25.8 &  6.7 & {\bfseries 15.3} & {\bfseries 53.8} & {\bfseries 42.4} & 11.2 &    0 & 33.4 & 11.8 & 26.7\\ 
ours  & 71.5  & 80.8 & {\bfseries 61.0} & {\bfseries 21.1} & 92.2 & {\bfseries 33.5} & 6.8 & 14.8 & 48.3 & 42.3 & {\bfseries 13.2} & {\bfseries 13.9} & {\bfseries 35.3} & {\bfseries 13.2} & {\bfseries 30.4}\\ 

\hline
\end{tabular}
}

\caption{Results on the NYU dataset. Bold numbers represent the best scores.}
\vspace{-0.6cm}
\label{tab:QuantitativeResultsonNYU}
\end{center}
\end{table*}

%% file: figs/resNYUCAD.tex
\begin{table*}[t]
\begin{center}
\scalebox{0.91}
{
\begin{tabular} {l |c c c|c c c c c c c c c c c|c} \hline
 &  \multicolumn{3}{c|}{scene completion} & \multicolumn{12}{c}{semantic scene completion} \\ 
\hline
Methods  & prec. & recall & IoU & \cellcolor{rgb1}ceil. & \cellcolor{rgb2}floor & \cellcolor{rgb3}wall & \cellcolor{rgb4}win. & \cellcolor{rgb5}chair & \cellcolor{rgb6}bed & \cellcolor{rgb7}sofa & \cellcolor{rgb8}table & \cellcolor{rgb9}tvs & \cellcolor{rgb10}furn. & \cellcolor{rgb11}objs. & avg. \\ 
\hline
Zheng \etal~\cite{zheng2013beyond} 	& 60.1 & 46.7 & 34.6 & - & - & - & - & - & - & - & - & - & - & - & - \\ 
Firman \etal~\cite{firman2016NYUCAD} 	& 66.5 & 69.7 & 50.8 & - & - & - & - & - & - & - & - & - & - & - & - \\ 
SSCNet~\cite{song2017_SSCNet}   & 75.4 & {\bfseries 96.3} & 73.2 & 32.5 & 92.6 & 40.2 &  8.9 & 33.9 & 57.0 & {\bfseries 59.5} & 28.3 &  8.1 & {\bfseries 44.8} & 25.1 & 40.0\\ 
TS3D~\cite{Garbade2018_twoStream}      	& 80.2 & 91.0 & 74.2 & 33.8 & {\bfseries 92.9} & 46.8 & {\bfseries 27.0} & 27.9 & {\bfseries 61.6} & 51.6 & 27.6 & {\bfseries 26.9} & 44.5 & 22.0 & 42.1\\ 
ours        						& {\bfseries 88.7} & 88.5 & {\bfseries 79.4} & {\bfseries 54.1} & 91.5 & {\bfseries 56.4} & 14.9 & {\bfseries 37.0} & 55.7 & 51.0 & {\bfseries 28.8} & 9.2 & 44.1 & {\bfseries 27.8} & {\bfseries 42.8} \\
\hline
\end{tabular}
}  
\caption{Results on the NYUCAD dataset. Bold numbers represent the best scores.}
\vspace{-0.4cm}
\label{tab:QuantitativeResultsonNYUCAD}
\end{center}
\end{table*}

%% file: figs/visRes.tex
\begin{figure*}[t]
\begin{center}
{\includegraphics[width=0.95\linewidth]{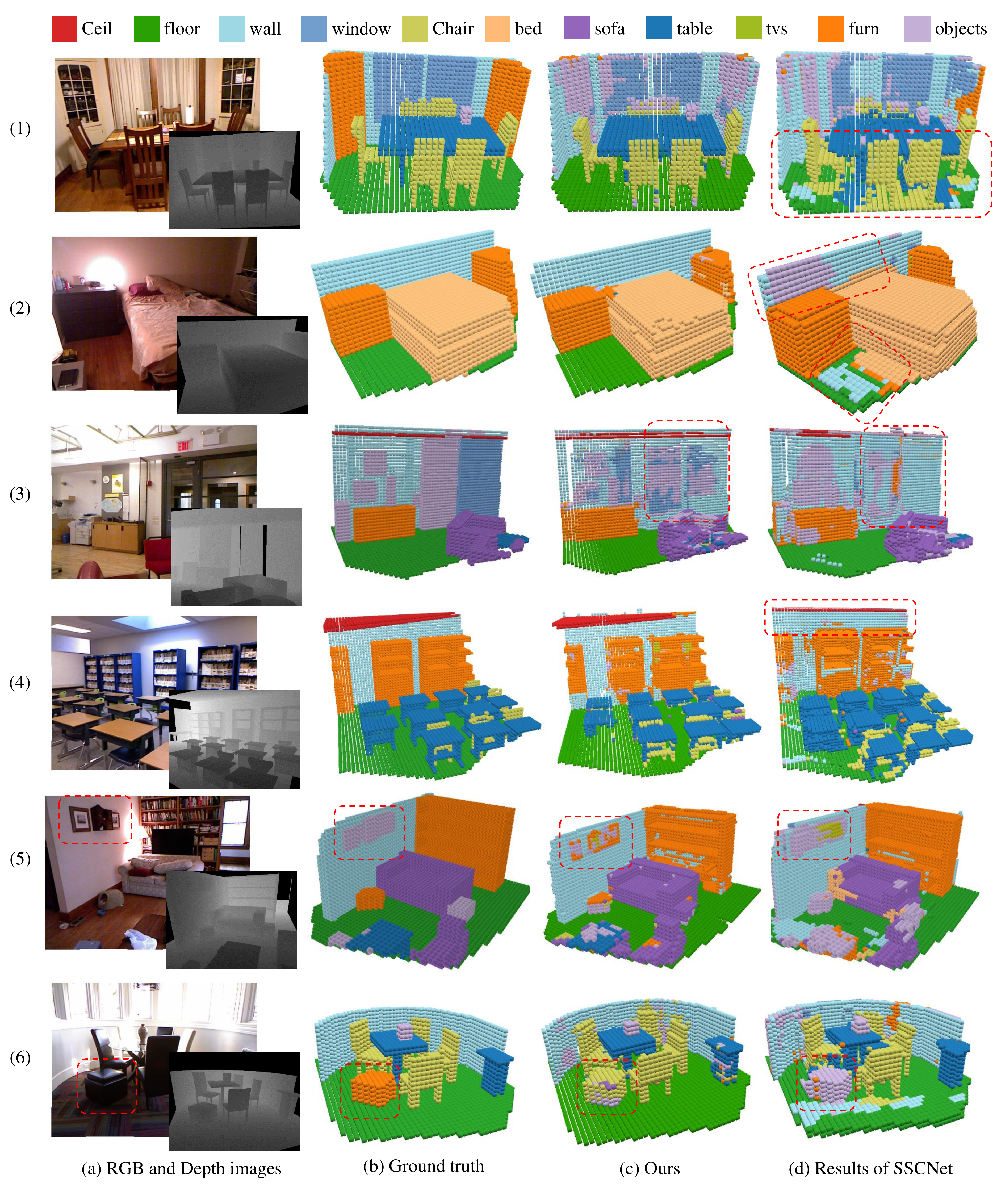}}

\caption{Qualitative results on NYUCAD. From left to right: Input RGB-D image,
ground truth, results obtained by our approach, and results obtained by SSCNet~\cite{song2017_SSCNet}. 
Overall, our completed semantic 3D scenes are less cluttered and show a higher voxel class accuracy compared to SSCNet. Refer to Section~\ref{qualitativeanalysis} for the detailed analysis.}
\end{center}
\label{fig:viz}
\end{figure*}

%% file: figs/rgbdepth.tex
\begin{table*}[t]
\begin{center}
\scalebox{0.91}
{
\begin{tabular} {l|c|c| c c c c c c c c c c c c} 
\hline
Methods 	    & Params/k 	& FLOPs/G & \cellcolor{rgb1}ceil. & \cellcolor{rgb2}floor & \cellcolor{rgb3}wall & \cellcolor{rgb4}win. & \cellcolor{rgb5}chair & \cellcolor{rgb6}bed & \cellcolor{rgb7}sofa & \cellcolor{rgb8}table & \cellcolor{rgb9}tvs & \cellcolor{rgb10}furn. & \cellcolor{rgb11}objs. & avg.\\ 
\hline
Ours-Depth 	& 155.0  	& 20.6  & 30.6 & 93.0 & 28.6 & 6.7 & 13.6 & 60.3 & 20.0 & 12.3 &  0.  & 30.9 & 12.0 & 28.9\\ 
Ours-RGB 	& 155.0  	& 20.6  & 19.3 & 91.8 & 30.5 & 3.7 & 13.1 & 44.4 & 37.1 & 10.6 & 5.5  & 31.0 & 11.9 & 27.2\\ 
Ours-RGBD 	& 195.0  	& 27.2  & 21.1 & 92.2 & 33.5 & 6.8 & 14.8 & 48.3 & 42.3 & 13.2 & 13.9 & 35.3 & 13.2 & 30.4\\
\hline
\end{tabular}
}
\caption{Ablation experiments of RGB and Depth fusion.}
\label{tab:ablationrgbdepth}
\vspace{-0.5cm}
\end{center}
\end{table*}

%% file: figs/resQuantitative.tex
\begin{table}[t]
\begin{center}
\scalebox{0.9}
{
\begin{tabular} {l|c|c|c|c} 
\hline
Methods 									& Params/k 	& FLOPs/G 		& SC-IoU	& SSC-IoU \\ \hline
SSCNet~\cite{song2017_SSCNet} 			& 930.0  	& 163.8  		& 55.1		& 24.7 \\   
EsscNet~\cite{zhang2018efficient}    	&  - 		& 22.0  			& 56.2		& 26.7 \\
Ours-Depth 								& 155.0  	& 20.6  		& 59.0		& 28.9 \\ 
Ours-RGBD 								& 195.0  	& 27.2  		& 61.0		& 30.4 \\ 
\hline
\end{tabular}
}
\caption{Params, FLOPs and Performance of our approach compared with other methods.}
\vspace{-0.6cm}
\label{tab:ParamsFLOPsofDifferent}
\end{center}
\end{table}

%% file: figs/paramsASPP.tex
\begin{table}[t]
\begin{center}
\scalebox{0.85}
{
\begin{tabular} {l|c|c|c|c} 
\hline
Method 	    		& Params/k 	& FLOPs/G	& SC-IoU	& SSC-IoU\\ \hline
Without ASPP	& 132.0  	&  21.13	& 56.8		& 26.8 \\ 
3D-ASPP	& 431.0 	&  63.28	& 62.7 		& 30.8 \\
LW-ASPP	& 195.0  	&  27.22  	& 61.0		& 30.4 \\ 
\hline
\end{tabular}

}

\caption{Params, FLOPs and Performance with/without ASPP.}
\label{tab:paramsASPP}
\end{center}
\vspace{-0.6cm}
\end{table}


%% file: figs/tab_MemSpeed.tex
\begin{table}[t]
\begin{center}
\scalebox{0.7}
{
\begin{tabular} {l|c|c|c|c|c|c} 
\hline
\multicolumn{1}{c|}{Method} & \multicolumn{1}{c|}{\begin{tabular}[c]{@{}c@{}}Params\\ (k)\end{tabular}} & \multicolumn{1}{c|}{\begin{tabular}[c]{@{}c@{}}FLOPs\\ (G)\end{tabular}} & \multicolumn{1}{c|}{\begin{tabular}[c]{@{}c@{}}Speed\\ (FPS)\end{tabular}} & \multicolumn{1}{c|}{\begin{tabular}[c]{@{}c@{}}Memory\\ (M)\end{tabular}} & \multicolumn{1}{c|}{\begin{tabular}[c]{@{}c@{}}Network \\ Depth\end{tabular}} & \multicolumn{1}{c}{\begin{tabular}[c]{@{}c@{}}SSC-IoU\\ (\%)\end{tabular}}         \\  \hline
SSCNet~[34] 	& 930.0             & 163.8         & 0.7          & 5305          & 14             & 24.7          \\
Ours-3D-ResNet  & 1540.5            & 204.7  	    & 1.3          & 1841          & 28             & \textbf{30.8} \\
Ours-DDR        & \textbf{195.0}    & \textbf{27.2} & \textbf{1.5} & \textbf{1829} & \textbf{44}    & 30.4          \\
\hline
\end{tabular}
}
\caption{The inference speed and GPU memory usage of our DDRNet and the 3D-ResNet based networks. All results are acquired on a GTX1080ti GPU and evaluated on the NYU[33] test set. }
\label{tab:SpeedMemory}
\end{center}

\vspace{-0.6cm}
\end{table}